\documentclass[runningheads]{llncs}
\usepackage[T1]{fontenc}
\usepackage{graphicx}
\usepackage{booktabs}  % 提供了 \toprule, \midrule, \bottomrule 等命令

\usepackage{amsmath}   % 更好的数学公式支持
\usepackage{float}     % 允许使用 [H] 强制固定图片或表格位置

\usepackage{tabularx}
\usepackage{booktabs}

\usepackage{rotating}
\usepackage{amssymb}
\usepackage{listings} % For code snippets
\usepackage{xcolor}
\usepackage{float}
\usepackage{caption}
\usepackage{algorithm}
\usepackage{algorithmic}
\usepackage{multirow}
\usepackage{longtable}
\usepackage{fancyvrb}
\usepackage{authblk} % For author affiliation
\usepackage{url}
\usepackage{marvosym}

\begin{document}

\title{Structure-BiEval: A Self-Supervised, Dual-Track Framework for Decoupling Structure and Content in LLM Evaluation for Web Information Systems}
\titlerunning{Structure-BiEval}
% If the paper title is too long for the running head, you can set
% an abbreviated paper title here
%
\iffalse
\author{Anonymous Author}
\institute{Affiliation\\
\email{email@example.com}\\
}
\fi

\author{Boxiang Zhao\textsuperscript{(\Letter)} \and Qince Li \and Zhonghao Wang \and Zelin Cao \and Yi Wang   \and \mbox{Peng Cheng  \and Bo Lin}}
%\author{Boxiang Zhao\textsuperscript{(\Letter)} \and Qince Li  \and Zhonghao Wang}
%\author{Authors}
%
\authorrunning{B. Zhao et al.}
% First names are abbreviated in the running head.
% If there are more than two authors, 'et al.' is used.
%
\institute{Tele-Communication Technology Bureau, Xinhua News Agency\\
\email{zhaobx9676@gmail.com}%\\\email{\{zhaoboxiang,liqince,wangyi08,zhonghaowang,chengpeng,linbo\}@xinhua.org}
}

\maketitle              % typeset the header of the contribution
\begin{abstract}
As Large Language Models (LLMs) evolve into the core of Web-based autonomous agents and complex Web Information Systems, their ability to faithfully translate natural language into rigorous structured formats has become paramount, as this capability is critical for Web API invocation and data exchange. However, evaluating this structural fidelity in Web-native payloads remains a challenge: traditional text metrics fail to capture topological consistency in semi-structured Web data, while manual evaluation is prohibitively costly. To address this, we propose Structure-BiEval, a novel self-supervised framework for quantitative, annotation-free assessment tailored for Web data engineering. By leveraging deterministic Intermediate Representations, our framework effectively decouples structure from content, utilizing Content Semantic Accuracy and Normalized Tree Edit Distance as precise metrics. We empirically benchmark 15 state-of-the-art LLMs across dual Web structural topologies, namely Hierarchical Data (Web backend payloads) and Tabular Data (Web frontend presentation). The results reveal substantial variability in structural performance, including cases where mid-sized models unexpectedly outperform larger counterparts in Web data formatting. Furthermore, our findings show that deep recursive nesting poses a consistent challenge for Web agents across varying \mbox{parameter scales}.

\end{abstract}
%
% ---------------------------------------------------------
% Section 1: Introduction
% ---------------------------------------------------------
\section{Introduction}
\label{sec:intro}

The role of Large Language Models (LLMs) is undergoing a paradigm shift from conversational interfaces to the backbone of autonomous agents and Retrieval-Augmented Generation (RAG) workflows~\cite{DBLP:journals/chinaf/XiCGHDHZWJZZFWXZWJZLYDW25}. Unlike early models evaluated primarily on linguistic fluency, these systems are now deeply embedded in software pipelines, necessitating the generation of rigorous, machine-readable data alongside human-readable text~\cite{DBLP:conf/iclr/YaoZYDSN023}. This transition imposes a dual requirement: mastery of data exchange standards for tool invocation and knowledge structuring for information presentation. Consequently, Text-to-Structure capability—the ability to faithfully translate unstructured intent into precise, executable heterogeneous data formats—has emerged as a critical competency for assessing the industrial viability of LLMs.

Despite the impressive performance of State-of-the-Art models in code generation~\cite{roziere2023code,DBLP:conf/acl/ChenDDZWWT0025}, practical deployment reveals significant unreliability when models process specific data topologies. We identify that this instability is not uniformly distributed but manifests in two distinct error patterns dictated by the topological characteristics of the data. First, regarding hierarchical data, models frequently succumb to contextual loss as nesting depth increases, reflecting a fundamental inability to manage recursive logic and long-range dependencies~\cite{DBLP:conf/iclr/HuangYBSKZN025}. Second, concerning tabular data, models face the challenge of spatial alignment in grid-like structures, where errors typically stem from a deficiency in linearizing two-dimensional information~\cite{wu2022text}.

Compounding this issue is the inadequacy of existing evaluation methodologies. Current frameworks predominantly rely on extensive human verification to judge semantic correctness, a process that is prohibitively expensive and inherently non-reproducible. While automated n-gram metrics like BLEU exist, they are notoriously insensitive to the semantic drift and structural nuances of structured data~\cite{DBLP:conf/eacl/LorandiB24}, rendering them unsuitable for industrial-grade assessment. Furthermore, these general-purpose metrics lack the granularity to distinguish between semantic reasoning errors and syntactic formatting failures, limiting their utility for targeted model optimization.

To reconcile the conflict between evaluation accuracy and cost, this work draws inspiration from the principles of reverse engineering~\cite{DBLP:conf/nips/HeXQWYLM16} to propose a fully automated, self-supervised evaluation paradigm. Our core hypothesis is that true understanding implies reversible reconstruction~\cite{DBLP:conf/iccv/ZhuPIE17}. If a model genuinely masters the topological structure and semantic content of data, it should be able to preserve information integrity during the bidirectional transformation between Data-to-Description and Description-to-Data.

Building on this premise, we introduce the Structure-BiEval framework. This framework eliminates the need for human annotation by establishing two automated evaluation tracks via deterministic algorithms: (1) Hierarchical Eval, which focuses on JSON and XML formats to rigorously test the restoration of tree structures in complex nested environments; and (2) Tabular Eval, which covers CSV, HTML, Markdown, LaTeX, and flattened JSON/XML lists, assessing the consistency of constructing two-dimensional data across multiple serialization forms. By incorporating an Intermediate Representation (IR), we parse heterogeneous raw data into a unified logical structure, upon which we calculate automated metrics—Content Semantic Accuracy (CSA) and Normalized Tree Edit Distance (NTED)—thereby achieving a zero-human-intervention metric of model capabilities.

The main contributions of this paper are summarized as follows:

\begin{itemize}
    \item \textbf{Structure-BiEval Framework:} We propose an automated, self-supervised evaluation framework for structured generation in Web Information Systems. By adopting an IR-based loop, it eliminates manual annotation and enables the scalable assessment of Web agents.

    \item \textbf{Web-Centric Dual-Track Benchmark:} We construct a comprehensive benchmark to separately evaluate Hierarchical Data (Web backend payloads) and Tabular Data (Web frontend presentation), enabling systematic analysis across diverse Web-native formats.

    \item \textbf{Structure-Aware Metrics:} We introduce two metrics, NTED and CSA, to quantitatively assess topological correctness and semantic fidelity. Unlike n-gram metrics, they are specifically designed to capture structural misalignment in semi-structured Web data.
\end{itemize}

% ---------------------------------------------------------
% Section 2: Related Work
% ---------------------------------------------------------
\section{Related Work} \label{sec:related_work} 
We contextualize our contributions by surveying the landscape of structured data processing, categorizing prior research into extraction tasks, executable semantics, code generation, and the evolution of evaluation metrics within the broader context of Web Information Systems.

\subsubsection{\textbf{Structured Data Understanding and Information Extraction}} Early research in structured data understanding primarily focused on extracting atomic entities using discriminative models~\cite{lample2016neural,devlin2018bert}. With the rise of Large Language Models, the paradigm has shifted towards generative extraction, where models synthesize holistic structures rather than merely tagging tokens. Wu et al. introduced the Text-to-Table task~\cite{wu2022text} for tabular summarization, while the demand for Web-based autonomous agents has spurred benchmarks like ToolBench~\cite{DBLP:conf/iclr/QinLYZYLLCTQZHT24} and Gorilla~\cite{DBLP:conf/nips/PatilZ0G24}. These benchmarks evaluate the generation of executable JSON arguments, which are the foundational data exchange formats for Web API invocation.

Despite these advancements, existing approaches face limitations regarding topological scope and evaluation granularity. Text-to-Table is confined to flat grid topologies, overlooking the recursive logic inherent in hierarchical data. Conversely, agent-oriented benchmarks predominantly prioritize execution success rates, often neglecting the intrinsic structural fidelity of the generated artifacts. Unlike this study, which systematically assesses heterogeneous topologies across a multi-dimensional spectrum, prior work lacks a universal, format-agnostic metric for quantifying the structural integrity of generated Web data outputs.

\subsubsection{\textbf{Text-to-SQL and Executable Semantic Evaluation}} Executable semantics represent a highly reliable evaluation methodology for structure generation, particularly for the backend databases driving Web applications. Text-to-SQL tasks avoid the superficial consistency issues of pure text matching by comparing the execution results of generated SQL queries against standard answers~\cite{DBLP:journals/csur/ShiTZZY26,DBLP:journals/vldb/KatsogiannisMeimarakisK23}. The Spider dataset~\cite{yu2018spider} established semantic execution as a mainstream metric by introducing multi-table nesting. More recently, the BIRD benchmark~\cite{li2024can} has pushed the boundaries of this paradigm, evaluating LLMs on massive, real-world databases to assess complex reasoning capabilities.

Yet, these approaches are strictly bound to query languages with explicit execution environments. For generic structured sequence generation across diverse Web services, there is no corresponding execution engine to serve as an oracle. Consequently, this study proposes an IR-based closed-loop framework to circumvent the dependency on execution semantics, thereby extending rigorous evaluation to cover all mainstream Web serialization formats.

\subsubsection{\textbf{Code Generation and Program Repair}} Code generation tasks share substantial methodological parallels with structured data generation, as both require strict adherence to grammatical constraints~\cite{DBLP:conf/iclr/ZhuoVCH0WYZHPB025}. While early models established capabilities on snippet-level tasks, recent benchmarks like SWE-bench~\cite{jimenez2024swebench} have shifted focus to repository-level software engineering, revealing that models often fail to maintain logical consistency across long contexts despite generating syntactically valid functions. In the realm of error correction~\cite{DBLP:conf/icse/FanGMRT23}, the paradigm has evolved from static analysis to dynamic iterative refinement. Advanced frameworks like Reflexion~\cite{shinn2024reflexion} enable models to self-repair by leveraging execution feedback and unit test results as verbal reinforcement signals during the iterative process.

Crucially, however, these methods rely fundamentally on the existence of an executable environment and a predefined test suite. They offer no solution for evaluating or repairing inert structured data where "execution" is undefined, highlighting the necessity for our intrinsic structural evaluation metrics to bridge this methodological gap in Web data engineering.

\subsubsection{\textbf{Structured Generation in Large Language Models}} Recent literature has begun to recognize the paradigmatic divergence between structured generation and natural language generation. Ouyang et al. noted in InstructGPT~\cite{ouyang2022training} that alignment training often biases models towards generating user-friendly text rather than the strict, machine-readable structures required by Web protocols. Addressing this, Zhuang et al. introduced StructLM~\cite{zhuang2024structlm}, demonstrating that specialized instruction tuning on structured data significantly enhances the ability of models to manipulate tables and hierarchical Web-native formats. Furthermore, the focus has shifted towards complex constraint satisfaction. Wen et al. proposed ComplexBench~\cite{wen2024benchmarking}, evaluating the capacity of LLMs to follow instructions with composed constraints, including strict formatting rules.

While these benchmarks assess whether a model attempts to satisfy constraints, they primarily rely on rule-based parsing or keyword matching. They often lack a fine-grained metric to quantify the degree of structural preservation when generation fails. Unlike prior approaches that focus on Abstract Syntax Tree (AST) for code~\cite{jiang2022an} or grammar-based decoding~\cite{yin2017syntactic}, our work provides a universal, post-hoc evaluation framework that measures the semantic drift in heterogeneous Web data formats without imposing external decoding constraints during the generation process.

\subsubsection{\textbf{Evaluation Metrics for Structured Outputs}} Traditional N-gram metrics such as BLEU~\cite{papineni2002bleu} are widely used but inadequate for capturing the topological nuances of semi-structured Web data. While Tree Edit Distance algorithms~\cite{pawlik2015efficient} address structural differences, they have historically lacked integration into generative evaluation pipelines. Recently, the paradigm has shifted towards LLM-as-a-judge frameworks, exemplified by G-Eval~\cite{liu2023geval} and Prometheus 2~\cite{kim2024prometheus}, which employ strong models to score outputs~\cite{DBLP:conf/acl/0001GM25}.

Although these methods align well with human perception, they suffer from non-determinism, high latency, and self-preference bias, making them difficult to scale within automated Web Information Systems. Unlike these probabilistic approaches, our work revisits rigorous mathematical foundations, introducing CSA and NTED. These metrics utilize semantic triples and tree algorithms to establish a deterministic, computationally efficient standard for quantifying both content fidelity and structural integrity.

\begin{figure}[t]
    \centering
    \vspace{-1mm}
    \includegraphics[width=0.9\linewidth]{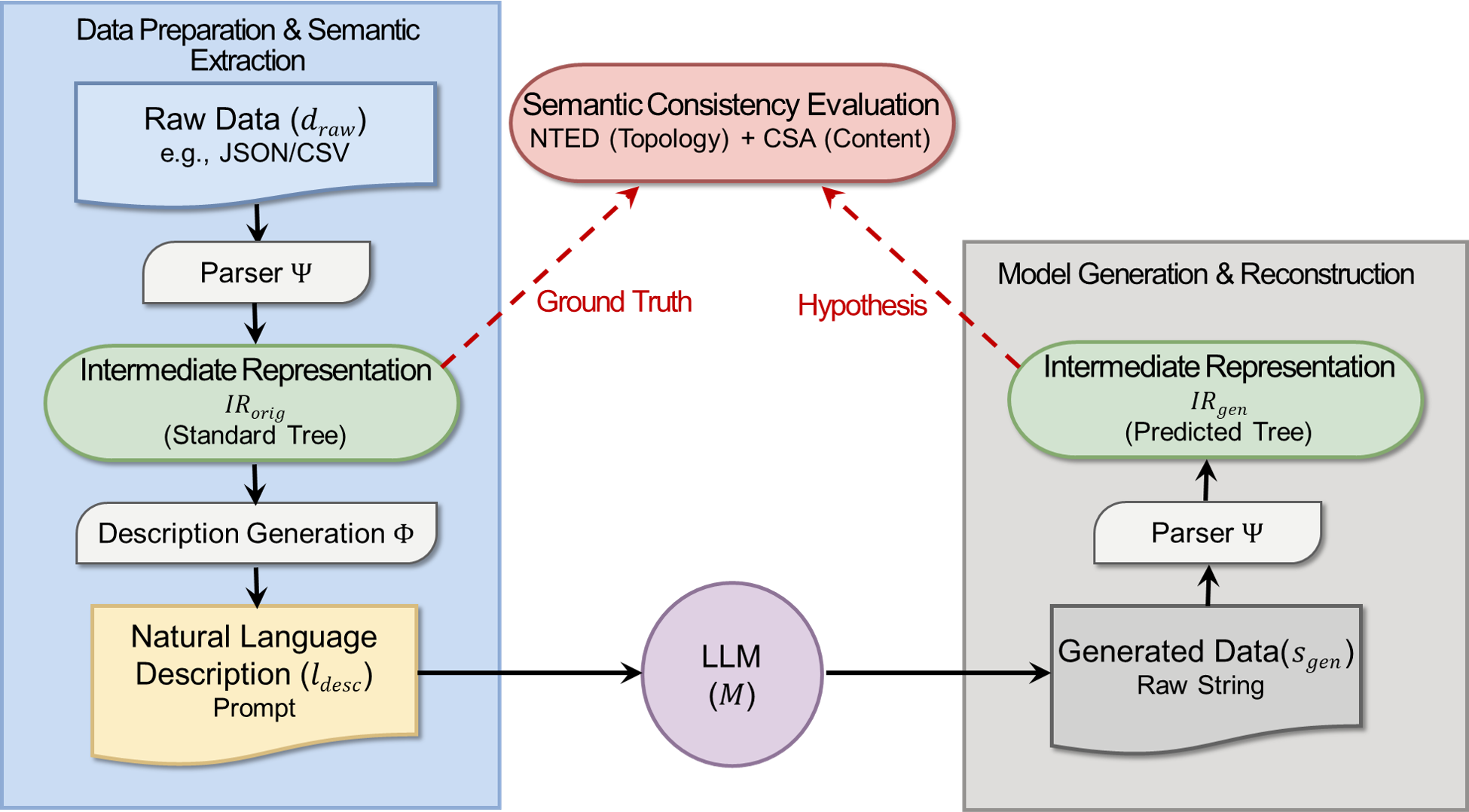}
    \vspace{-1mm}
\caption{Overview of the Structure-BiEval Framework. %The pipeline operates as a five-stage self-supervised closed loop. First, heterogeneous raw data is parsed into a unified Intermediate Representation ($IR_{orig}$). Second, a prompt generator converts the IR into a natural language description to simulate user intent. Third, the target LLM generates structured outputs based on this description. Fourth, the output is re-parsed into $IR_{gen}$. Finally, the system computes Topological Consistency (via NTED) and Atomic Fidelity (via CSA) by aligning the two IRs, enabling fully automated evaluation without human annotation.
}    \label{fig:framework}
\vspace{-3mm}
\end{figure}

% ---------------------------------------------------------
% Section 3: Structure-BiEval Framework
% ---------------------------------------------------------
\section{Structure-BiEval Framework}
\label{sec:framework}

This section details the design of the evaluation framework. As illustrated in Figure~\ref{fig:framework}, Structure-BiEval departs from traditional textual comparison methods to establish a five-stage closed-loop validation process based on IR: Raw Data $\to$ $IR_{orig}$ $\to$ Description $\to$ Generated Data $\to$ $IR_{gen}$ $\to$ Semantic Comparison.

\subsection{Stage 1: Intermediate Representation Abstraction}
To eliminate the syntactic noise introduced by heterogeneous Web serialization formats—such as delimiters in CSV or enclosing tags in XML—we establish a unified internal structure space $\mathcal{I}$, implemented as a format-agnostic AST. This design aims to encapsulate the pure topological characteristics of data, decoupling semantic logic from surface syntax.

Let $d_{raw} \in \mathcal{D}$ denote a raw data sample from the diverse Web-native format set $\{JSON, XML, CSV, \dots\}$. We define a deterministic parsing function $\Psi: \mathcal{D} \to \mathcal{I}$ to map raw files into this abstract space:
\begin{equation}
    IR_{orig} = \Psi(d_{raw})
\end{equation}
The mapping process $\Psi$ performs structural normalization through two specialized logical paths. For hierarchical data (predominant in Web backend communications), entities are recursively parsed into a tree of typed nodes (\texttt{DICT}, \texttt{LIST}, \texttt{VALUE}), preserving nested dependencies. Conversely, for tabular data (common in frontend Web presentations), the parser reconstructs the grid topology into semantic components (\texttt{Header}, \texttt{Row}, \texttt{Cell}), thereby unifying diverse delimiters into a consistent object model. This transformation ensures that $IR_{orig}$ serves as the unbiased semantic ground truth for subsequent evaluation. Detailed definitions of the AST node structures are provided in Appendix A.

\subsection{Stage 2: IR-Based Description Generation}
Different structured data types correspond to distinct cognitive patterns. Grid-based formats align with row/column spatial thinking, whereas hierarchical formats align with parent/child recursive thinking.

To simulate real-world Web interaction scenarios, we design a description generation function $\Phi: \mathcal{I} \times \mathcal{T} \to \mathcal{L}$, where $\mathcal{T}$ represents the format type of the raw data. Based on $IR_{orig}$, we generate a natural language description $l_{desc}$ that encapsulates both data content and implicit topological characteristics.

For grid-based formats, $\Phi$ employs a coordinate-based description strategy. It iterates through the rows and columns of $IR_{orig}$, generating instructions such as "In row $i$, the value of column $h$ is $v$," effectively simulating the logic a user employs when reading report data or frontend Web tables. Conversely, for tree-based formats, $\Phi$ utilizes a path-based description strategy. By performing a Depth-First Search traversal of the $IR_{orig}$ tree structure, it generates instructions such as "Under the path root/users/id, there exists a node with value 101," thereby simulating the logic used in defining hierarchical Web API payloads. By adopting this approach, we minimize ambiguity during the intent understanding process, ensuring that the Input Prompt contains the necessary and sufficient information required to reconstruct $IR_{orig}$. To ensure the generative task is not artificially simplified, we strictly control the ``hint strength'' of the natural language description. The generator $\Phi$ operates in a schema-agnostic manner, employing neutral templates without leaking any serialization-specific syntax (e.g., omitting curly braces for JSON, or angular brackets for XML). This guarantees that the model must rely on its intrinsic structural reasoning to construct the target topology, rather than superficially copying syntax hints from the prompt. See Appendix B for details.

\subsection{Stage 3: Model Generation and IR Remapping}
In this phase, the Large Language Model $M$, acting as a Web agent, receives the description $l_{desc}$ along with target format instructions and outputs a generated string sequence $s_{gen}$. Subsequently, we employ the parsing function $\Psi$ once again to map the generated string back into the IR space:
\begin{equation}
    IR_{gen} = \Psi(s_{gen})
\end{equation}
If $s_{gen}$ contains severe syntactic errors that prevent parsing (like missing closing brackets in JSON), then $IR_{gen} = \bot$ (the empty set). This step anchors the unstructured generated text back into the structured logical space.

We define a hard invalid output ($IR_{gen} = \bot$, resulting in $\text{CSA} = 0$ and $\text{NTED} = 0$) strictly as cases where the extracted payload fails the standard deterministic parser. We deliberately do not employ LLM-based self-correction or heuristic structural guessing to fix broken brackets, ensuring that our metrics objectively reflect the raw structural fidelity of the model's single-turn output, which is a critical requirement for low-latency Web applications.

\subsection{Stage 4: Semantic Metrics in Dual-IR Space}
To comprehensively assess model capabilities in terms of both structural maintenance and content generation, we define two complementary metrics within the IR space. Let $IR_{orig}$ denote the standard tree structure parsed from the raw data, and $IR_{gen}$ denote the predicted tree structure parsed from the model-generated data.

\subsubsection{Topological Consistency: Normalized Tree Edit Distance}
In nested data structures, even if all key-value pairs are correct, a misalignment in the nesting hierarchy—such as incorrectly mounting an address field under the root rather than the user node—fundamentally alters the semantics of the data. We adopt Tree Edit Distance (TED) to quantify this deviation in topological structure.

We treat $IR_{orig}$ and $IR_{gen}$ as Labeled Ordered Trees. The calculation involves three fundamental edit operations $\delta$: insertion, deletion, and renaming. Following standard practices in evaluating structured abstractions~\cite{DBLP:conf/acl/SongLTBK24}, we assign a uniform unit cost to all operations, i.e., $\gamma(\text{insert}) = \gamma(\text{delete}) = \gamma(\text{rename}) = 1$. While weighted costs can be tailored for specific downstream applications, unit costs provide a generalized, unbiased distance metric for topological structures independent of specific semantic payloads. The Tree Edit Distance is formulated as:
\begin{equation}
    \text{TED}(IR_{gen}, IR_{orig}) = \min_{\{s_1, \dots, s_k\}} \sum_{i=1}^{k} \gamma(s_i)
\end{equation}

Finally, to ensure fair comparison across samples of varying sizes, we normalize the metric into the $[0, 1]$ interval, where 1 indicates perfect structural identity:
\begin{equation}
    \text{NTED} = 1 - \frac{\text{TED}(IR_{gen}, IR_{orig})}{\max(|IR_{gen}|, |IR_{orig}|)}
\end{equation}
Here, $|IR|$ denotes the total number of nodes in the intermediate representation tree $IR$. This unifies the notation and avoids structural penalties from scaling strictly with sequence length.

\subsubsection{Atomic Fidelity: Content Semantic Accuracy}
Traditional metrics like BLEU and ROUGE fail to capture the semantics of structured data, as they are insensitive to numerical precision and heavily penalize the valid reordering of keys in unordered dictionaries. Content Semantic Accuracy (CSA) addresses this by evaluating data at the level of semantic triples. Unlike ROUGE, CSA is permutation-invariant, assessing each Atomic Information Unit based on its logical position and value rather than its sequential text appearance.

CSA computation relies on Semantic Flattening via a mapping function $F: \mathcal{I} \to \mathbf{\Omega}$. For every leaf node, we generate a triple $\omega = (p, k, v)$, where $p$ is the root-to-node path, $k$ is the field name, and $v$ is the type-normalized value to eliminate formatting noise. We flatten both data trees to $\Omega_{orig} = F(IR_{orig})$ and $\Omega_{gen} = F(IR_{gen})$, calculating CSA via the Jaccard similarity:
\begin{equation}
    \text{CSA} = \frac{|\Omega_{gen} \cap \Omega_{orig}|}{|\Omega_{gen} \cup \Omega_{orig}|}
\end{equation}
This formulation inherently penalizes hallucinations (increasing the denominator) and omissions (decreasing the numerator). Furthermore, its path-sensitivity ensures it effectively distinguishes between correct values placed in incorrect structural locations.

Crucially, the definition of path $p$ adapts to data formats. For hierarchical structures, $p$ is the sequence of keys. For tabular data, $p$ explicitly includes coordinate indices. While this order-sensitive formulation penalizes semantically valid row or column permutations, it is practically necessary for Web-based autonomous agents: downstream Web services and RESTful API invocations strictly expect schema positional alignment. Relaxing this would severely compromise programmatic indexing.

% ---------------------------------------------------------
% 第四章：实验评估 (Experimental Evaluation)
% ---------------------------------------------------------

\section{Experimental Evaluation}
\label{sec:evaluation}

To rigorously assess the capabilities of LLMs in generating structured data for modern Web Information Systems, we conducted a systematic evaluation using the \texttt{Structure-BiEval} benchmark. This section details our experimental setup, the models evaluated, and the metrics employed to quantify performance.

\subsection{Benchmark and Dataset}
Our evaluation utilizes the \texttt{Structure-BiEval} dataset, which is designed to probe model performance across two distinct topological dimensions representative of Web data exchange: Hierarchical Data  and Tabular Data. The dataset comprises 13 distinct categories, ranging from simple, flat records to complex, deeply nested structures and symbolic-heavy content. This diversity allows for a granular analysis of how different data characteristics—such as recursion depth, token verbosity, and escape character density—impact the generative fidelity of Web agents. Specifically, the dataset includes variations in core Web serialization formats (JSON, XML, HTML, CSV, Markdown, LaTeX) and complexity (Simple, Nested, Long-List, Sparse, Text-Heavy, Symbolic), ensuring a comprehensive stress test of the models' structural cognition in Web-centric environments.

\subsection{Models and Experimental Setup}
We evaluated a diverse suite of 15 state-of-the-art open-source LLMs, spanning a wide spectrum of parameter scales and architectural paradigms. The lineup includes the Qwen3 series (ranging from 0.6B to 235B parameters), the Llama-3.1 family (8B and 70B), and the DeepSeek suite (V3.1, V3.2, and the reasoning-enhanced R1). Additionally, we incorporated representative models including Seed-OSS-36B, Minimax-M1-80k, GPT-OSS-120B, and Kimi-k2-thinking, with the latter representing the trillion-parameter scale. 

All models were evaluated in a zero-shot setting to simulate real-world cold-start scenarios in autonomous Web agent deployments. To ensure reproducibility and isolate the models' inherent capabilities from prompt engineering variance, we employed a standardized system prompt that strictly defined the output format constraints without providing few-shot examples. The generation temperature was set to 0.1 to minimize stochasticity, reflecting the deterministic reliability required in Web systems engineering. Furthermore, a strict parsing pipeline was implemented to validate the syntactic correctness of the outputs before semantic evaluation, mirroring the strict payload validation processes of RESTful Web APIs.

\subsection{Evaluation Metrics}
Performance was quantified using two complementary metrics designed to decouple structural integrity from semantic accuracy. First, the NTED measures the topological isomorphism between the generated structure and the ground truth, capturing the model's ability to maintain the correct hierarchy and nesting logic independent of the leaf node content. Second, the CSA assesses the element-wise fidelity of the data, penalizing hallucinations or omissions in the actual values. By analyzing the divergence between these two metrics, we can effectively distinguish between models that genuinely understand the underlying Web data schema and those that merely mimic the syntactic surface form, a critical distinction for preventing silent failures and ensuring the robustness of downstream Web service integrations.

% ---------------------------------------------------------
% 第五章：实验结果与分析 (Results and Analysis)
% ---------------------------------------------------------
\section{Results and Analysis}
\label{sec:results}

In this section, we present a detailed analysis of the experimental results within the context of Web Information Systems. Our findings reveal complex non-linear relationships between model scale, architectural paradigm, and structured generation performance, challenging several prevailing assumptions about model capabilities in Web-centric environments.

\subsection{Performance Overview and Metric Sensitivity Analysis}
\label{subsec:performance_overview}

\begin{figure}[t]
    \centering

    \includegraphics[width=0.95\linewidth]{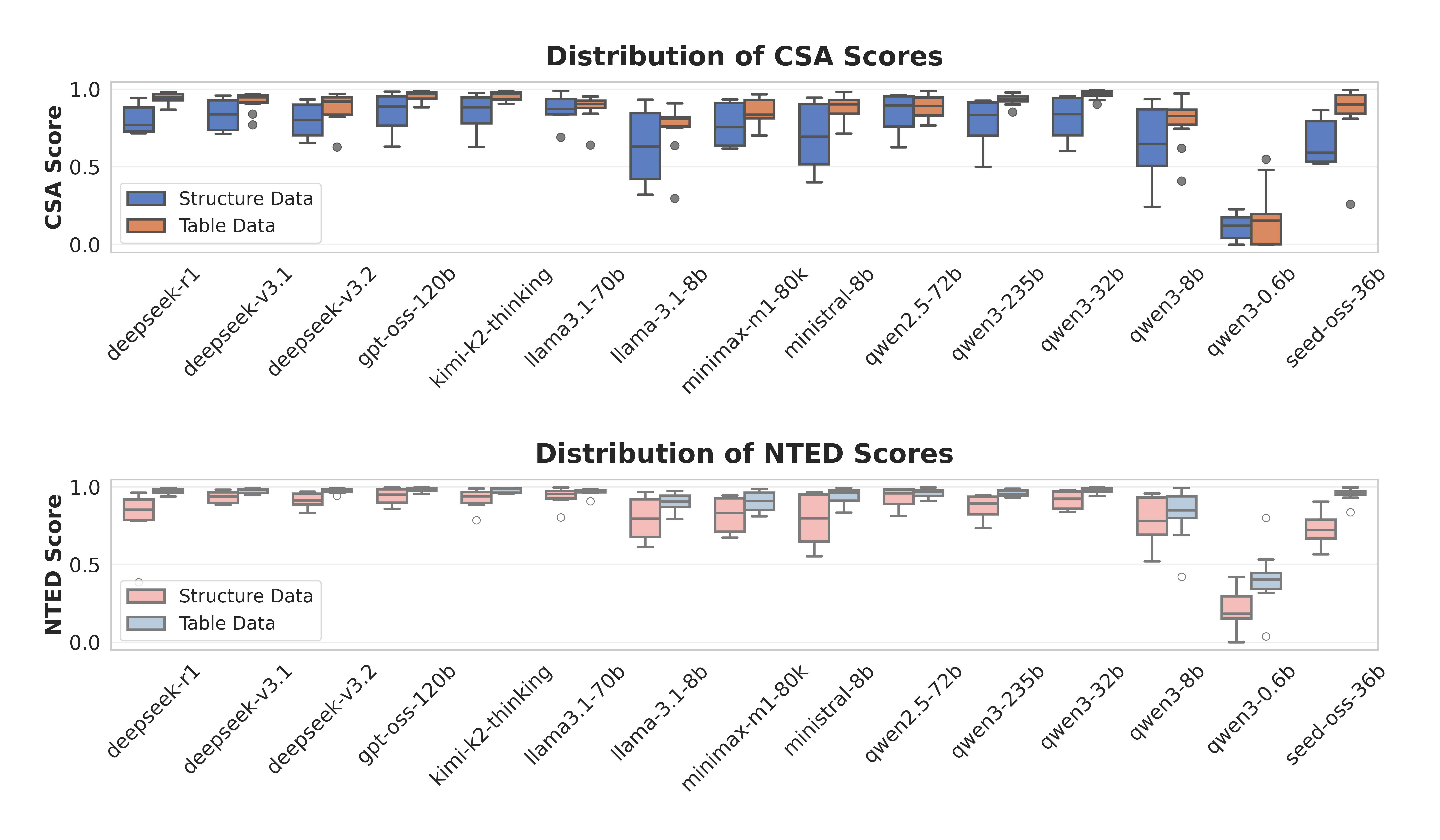}
    \vspace{-3mm}
    \caption{Stability Analysis. %Box plots reveal the distribution of CSA and NTED scores across models, with smaller models exhibiting larger variances and significant hierarchical-tabular gaps.
    }
    \label{fig:1_box_distribution}
    \vspace{-3mm}
\end{figure}

The overall performance landscape, as illustrated in Figure \ref{fig:2_overall_performance_bar} and Figure \ref{fig:1_box_distribution}, demonstrates a clear stratification of model capabilities. We explicitly incorporate ROUGE-2 as a baseline to highlight the comparative sensitivity of our proposed metrics. In terms of absolute fidelity, GPT-OSS-120B and DeepSeek-R1 lead the first tier, exhibiting exceptional robustness. Notably, the Qwen3-32B model delivers a standout performance, achieving a CSA score of 0.912 and an NTED score of 0.958. This mid-sized model not only outperforms its larger counterpart, Qwen3-235B (CSA 0.851), but also surpasses Llama-3.1-70B (CSA 0.877). This counter-intuitive result reinforces that for structured generation in Web-based workflows, the efficiency of instruction tuning and architectural optimization plays a more critical role than sheer parameter scaling.

\begin{figure}[t]
    \centering
    \vspace{-3mm}
    \includegraphics[width=0.95\linewidth]{2_overall_performance_ROUGE-2.png}
    \vspace{-3mm}
    \caption{Overall Performance Landscape and Metric Sensitivity. %The chart compares model performance across Hierarchical and Tabular tracks using our proposed metrics (CSA, NTED) against the ROUGE-2 baseline.
    }
    \label{fig:2_overall_performance_bar}
    \vspace{-3mm}
\end{figure}

A critical insight emerges from the variance analysis across different metrics. Traditional metrics exhibit a pronounced discriminative saturation, failing to effectively distinguish between models. For Hierarchical Data, the variance of our CSA metric ($\sigma^2=0.0601$) and NTED ($\sigma^2=0.0436$) is significantly higher than that of ROUGE-2 ($\sigma^2=0.0145$) and BLEU ($\sigma^2=0.0056$). A similar trend is observed in Tabular Data, where CSA variance ($0.0465$) is nearly triple that of ROUGE-2 ($0.0156$). 

This discrepancy reveals that traditional n-gram metrics tend to overestimate the capabilities of weaker models, creating an inflated perception of proficiency by rewarding surface-level token overlap in Web payloads. In contrast, our metrics expose the true performance gaps, effectively identifying the hollow structural generation where models mimic syntactic skeletons without maintaining semantic logic, which is a primary cause of critical parsing failures in Web API integrations. Comprehensive results for ROUGE-1 and BLEU, which exhibit similar low-variance behaviors, are detailed in Appendix C.

\begin{figure}[t]
    \centering
    \vspace{-3mm}
    \includegraphics[width=1.0\linewidth]{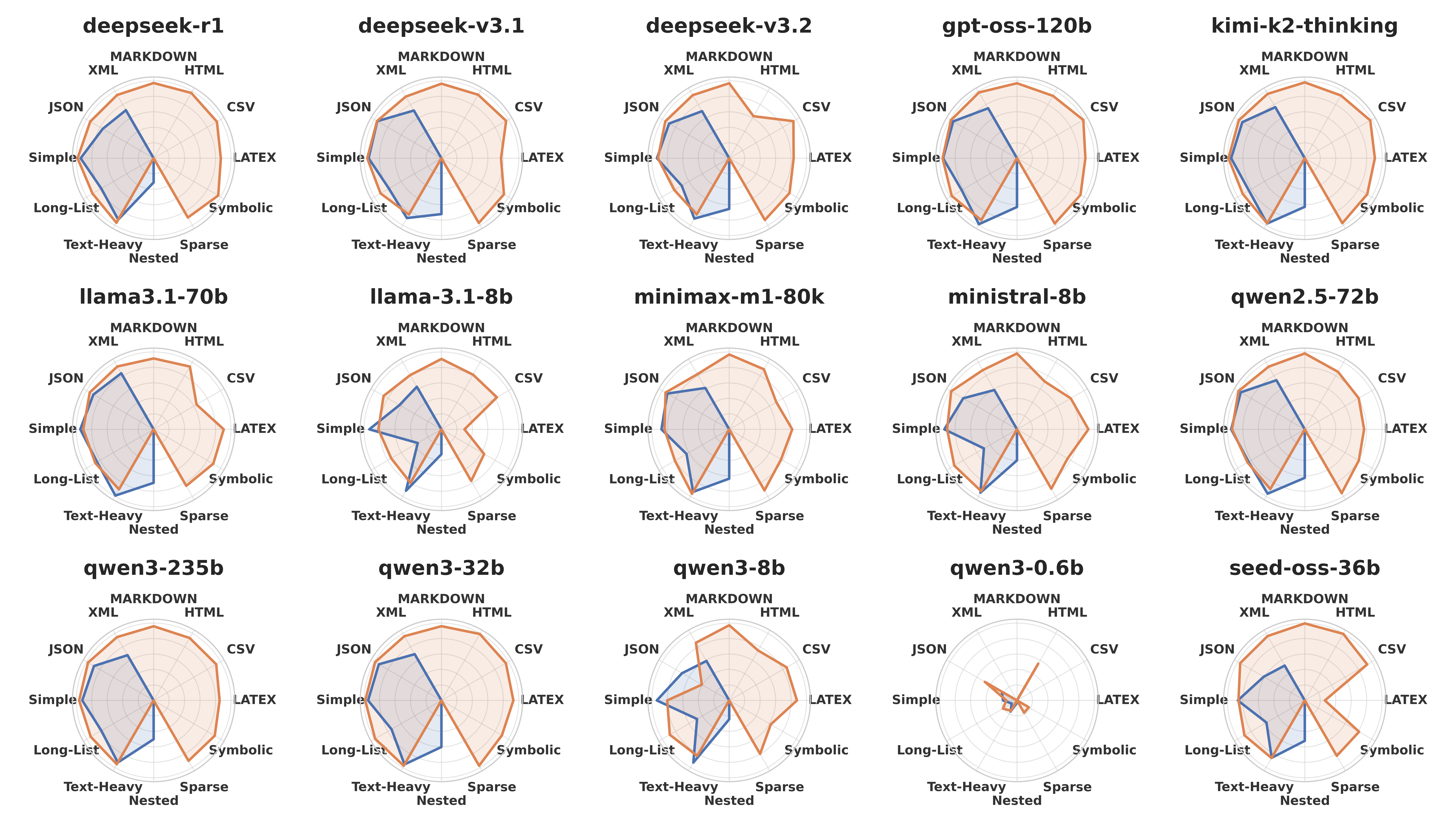}
    \vspace{-3mm}
    \caption{Format Adaptability Radar. %The radar chart illustrates the suppressive effect of HTML formats on smaller models and the universal advantage of JSON across most architectures in Tabular Data generation.
    }
    \label{fig:5_grid_radar_chart}
    \vspace{-3mm}
\end{figure}

\subsection{Topology Adaptability: The Dominance of JSON and the HTML Threshold}
\label{subsec:topology_adaptability}

Our evaluation of topological adaptability, summarized in Figure \ref{fig:5_grid_radar_chart}, challenges the traditional view that XML is superior for long-context generation. With the exception of Qwen3-8B, all other 14 models demonstrated significantly better performance on JSON formats compared to XML. Minimax-M1-80k, for instance, suffered a massive 21.7\% performance penalty on XML tasks, with DeepSeek-V3.2 showing a similar trend (10.3\% penalty). This pervasive dominance of JSON strongly supports the hypothesis that in the current era of pre-training dominated by code data, models have developed a stronger inductive bias towards JSON as a code-native format, making it a safer vector for modern Web API interactions than XML.

Furthermore, the evaluation of grid-based data reveals a significant performance divergence, particularly in the handling of HTML formats, which are foundational for Web frontend rendering. While Markdown serves as a robust baseline for all models with average CSA scores consistently above 0.95, the transition to the semantically equivalent but syntactically verbose HTML format triggers a drastic divergence. As expected, smaller models struggle with tag redundancy (e.g., Ministral-8B and Qwen3-8B experienced 26.6\% and 22.6\% drops, respectively, compared to Markdown). However, paradoxically, the massive MoE model DeepSeek-V3.2 experienced the most severe degradation—a 34.2\% drop. Conversely, dense architectures like Qwen3-32B (+3.2\%) and Llama-3.1-70B (+2.1\%) remained largely unaffected or even improved. This indicates that effectively suppressing tag noise and leveraging explicit HTML structure is not strictly a function of parameter scale; rather, it highlights a vulnerability in certain advanced alignment paradigms when processing verbose frontend presentation layers.

\begin{figure}[t]
    \centering
    \vspace{-3mm}
    \includegraphics[width=0.9\linewidth]{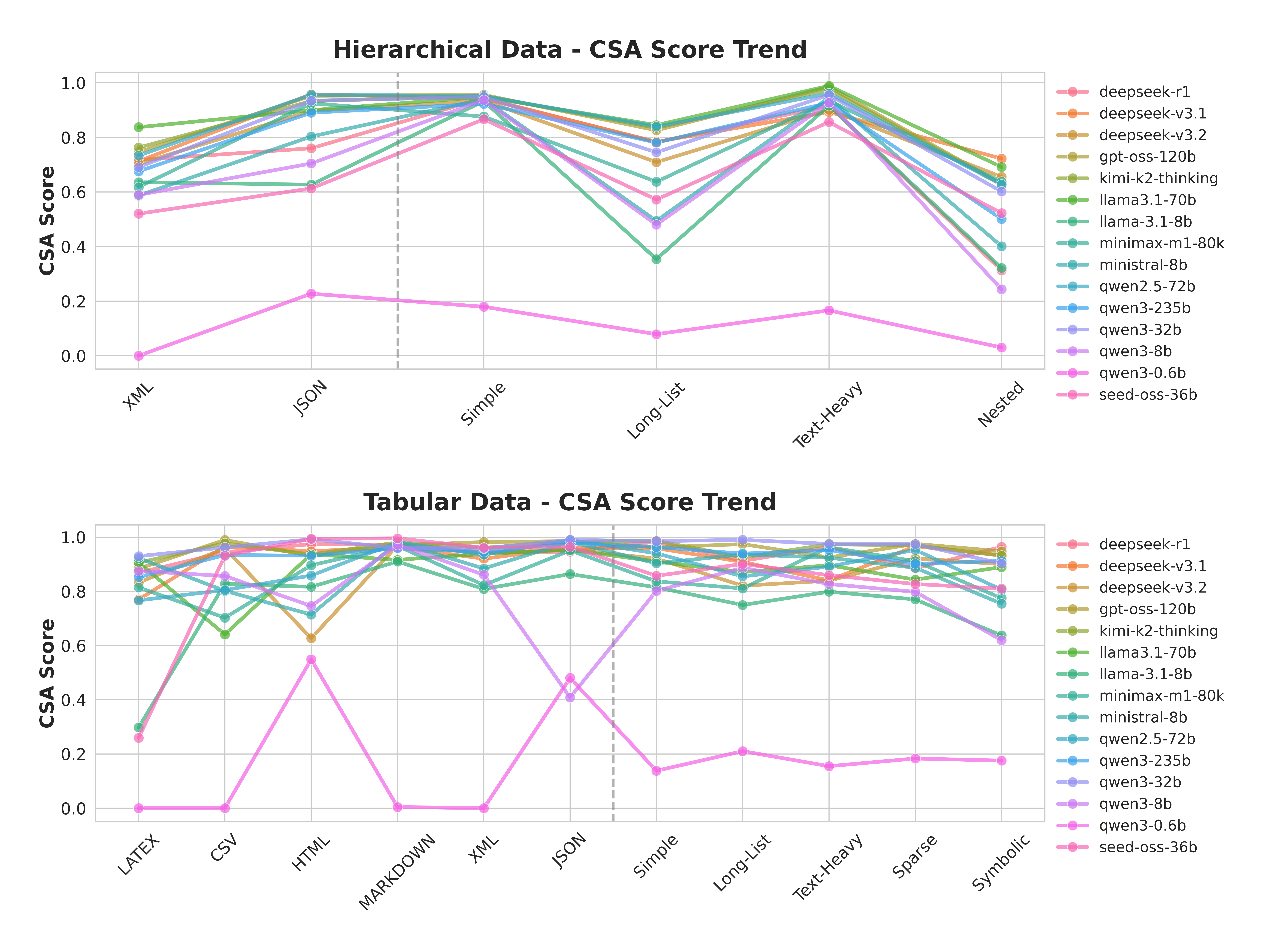}
    \vspace{-3mm}
    \caption{Performance Degradation Trend. %The sharp contrast between DeepSeek-R1's perfect performance on Symbolic tasks and its severe degradation on Nested tasks highlights the distinct cognitive demands of different data complexities in Hierarchical Data generation.
    }
    \label{fig:3_category_trend_line}
    \vspace{-3mm}
\end{figure}
\subsection{Complexity Robustness: The Trade-off Between Reasoning and Recursion}
\label{subsec:complexity_robustness}

The performance of models under extreme data complexity provides new insights into the cognitive boundaries of LLMs. As depicted in Figure \ref{fig:3_category_trend_line}, Nested tasks, analogous to deeply nested JSON responses in complex Web services, posed the most significant challenge for all models. When recursion depth exceeded three levels, performance across the board plummeted. Even the robust Llama-3.1-70B saw its CSA score drop from a baseline of 0.93 on Simple tasks to 0.69, highlighting that handling long-range recursive dependencies remains a fundamental intrinsic limitation for current Transformer architectures. Models struggle to maintain structural integrity and content accuracy simultaneously within deep contexts characterized by extensive nesting.

A more profound finding emerges from the analysis of DeepSeek-R1, which exhibits an instructive performance asymmetry. As a reasoning-enhanced model, DeepSeek-R1 achieved near-perfect performance on Symbolic tasks containing dense code symbols and escape characters, with almost zero degradation (-0.05\%), demonstrating exceptional mastery over formal language rules. However, its performance on Nested tasks collapsed by 67.6\%, a drop far exceeding that of the traditional DeepSeek-V3.1 architecture (\ensuremath{\sim}24\%). This phenomenon suggests that current Chain-of-Thought (CoT) training, while significantly enhancing linear logical reasoning, may not effectively improve and might even compromise the memory stack maintenance capabilities required for processing deep recursive structures. This underscores the need for a strategic trade-off when selecting models for tasks that prioritize logical reasoning versus topological fidelity in practical deployment scenarios for autonomous Web agents.

% ---------------------------------------------------------
% 第六章：结论 (Conclusion)
% ---------------------------------------------------------
\section{Conclusion}
\label{sec:conclusion}

This work presents Structure-BiEval, a self-supervised evaluation framework for assessing the structured Web data generation capabilities of LLMs within the context of Web Information Systems. By independently evaluating Hierarchical and Tabular topologies, and leveraging deterministic IR to compute CSA and NTED, Structure-BiEval offers a critical diagnostic perspective that traditional n-gram metrics lack, strictly focusing on structural fidelity in Web data exchange. An empirical study over 15 open-source models across different scales reveals notable variability in structural performance. Beyond parameter scale, instruction tuning tailored for Web-native formats emerges as a critical factor, with several mid-sized models achieving performance comparable to larger architectures. In addition, hollow structural generation in smaller models—a primary catalyst for silent parsing failures in Web API integrations—and the consistent performance degradation observed under deep recursive nesting point to a shared challenge in handling complex Web dependencies. The asymmetric behavior observed in reasoning-enhanced models further suggests that improvements in deductive reasoning do not necessarily translate into reliable topological consistency in complex Web structures.

These findings carry practical implications for the design of Web-based autonomous agents and resilient Web services. In modern Web engineering workflows, favoring JSON-style representations and flatter intermediate structures can significantly alleviate the contextual burden introduced by deep recursion during API communications. More broadly, the observed gap between linear text generation and hierarchical Web data construction highlights an important direction for future research in Web Intelligence. Structure-BiEval offers a reproducible benchmark to support systematic progress towards robust, Web-ready agentic intelligence.

% Bibliography entries for the entire Anthology, followed by custom entries
%\bibliography{anthology,custom}
% Custom bibliography entries only
\bibliographystyle{splncs04}
\bibliography{custom_cleaned.bib}

\clearpage
\appendix
\section{Detailed Definitions of Intermediate Representation}
\label{app:ir_definitions}

In Section \ref{sec:framework}, we established the Intermediate Representation (IR) as a format-agnostic Abstract Syntax Tree (AST). This appendix details the formal schema definitions and provides concrete instantiation examples to illustrate the parsing logic.

\subsection{Hierarchical Data AST (Hierarchical Track)}
For hierarchical data formats, the IR is modeled as a recursive tree structure. Formally, each node within the tree is instantiated as a 5-tuple $\mathcal{N}_{struct} = \langle T, K, V, I, \mathcal{C} \rangle$. The component $T \in \{ \texttt{ROOT}, \texttt{DICT}, \texttt{LIST}, \texttt{VALUE} \}$ designates the semantic category of the node, determining the validity of subsequent fields. The identifiers $K$ (string) and $I$ (integer) serve as addressing mechanisms for children within \texttt{DICT} and \texttt{LIST} nodes, respectively, whereas $V$ stores the primitive payload exclusively for \texttt{VALUE} nodes. The topological structure is maintained via $\mathcal{C}$, a sequence of child nodes allowing for arbitrary nesting depths.

\textbf{Illustrative Example:} 
Consider the raw JSON fragment \texttt{\{"user": ["Alice"]\}}. The parser transforms this into a three-level tree:
\begin{enumerate}
    \item \textbf{Level 0 (Root)}: $\langle \texttt{ROOT}, \emptyset, \emptyset, \emptyset, \{c_1\} \rangle$.
    \item \textbf{Level 1 (Dict Entry)}: The child $c_1$ is $\langle \texttt{DICT}, \text{"user"}, \emptyset, \emptyset, \{c_2\} \rangle$, representing the key.
    \item \textbf{Level 2 (List Container)}: The child $c_2$ is $\langle \texttt{LIST}, \emptyset, \emptyset, \emptyset, \{c_3\} \rangle$, representing the array value.
    \item \textbf{Level 3 (Leaf Value)}: The child $c_3$ is $\langle \texttt{VALUE}, \emptyset, \text{"Alice"}, 0, \emptyset \rangle$, where $0$ indicates the index.
\end{enumerate}
This abstraction strips away surface-level syntax (braces, brackets, colons) while preserving the topological path \texttt{root.user[0]}.

\subsection{Tabular Data AST (Tabular Track)}
For flat or grid-based serializations, the parsing logic normalizes diverse delimiter systems into a unified bipartite structure, defined as $\mathcal{IR}_{table} = \langle \mathcal{H}, \mathcal{R} \rangle$. Here, $\mathcal{H} = [h_1, h_2, \dots, h_n]$ represents the Header Node, a sequence of strings defining the semantic schema and column dimensionality of the table. The data body $\mathcal{R} = [r_1, r_2, \dots, r_m]$ constitutes a list of Row Objects, where each row $r_i$ is composed of a normalized vector of \texttt{CellNode} instances. This canonical form ensures that semantically identical tables are mathematically equivalent regardless of their serialization format.

\textbf{Illustrative Example:} 
Consider a Markdown table row \texttt{| Name | Age |} followed by \texttt{| Bob | 30 |}. The parser maps this to:
\begin{equation}
    \mathcal{IR}_{table} = \left\langle 
    \begin{aligned}
        \mathcal{H} &= [\text{"Name"}, \text{"Age"}] \\
        \mathcal{R} &= \left[ \langle \text{"Bob"}, \text{"30"} \rangle \right]
    \end{aligned}
    \right\rangle
\end{equation}
In this representation, the pipe delimiters (`|`) and spacing are discarded, retaining only the structural schema ($\mathcal{H}$) and the data payload ($\mathcal{R}$).

\section{Dataset and Prompting Details}
\label{app:dataset_prompts}

\subsection{Dataset Composition and Statistics}
To ensure a robust and comprehensive evaluation, the Structure-BiEval benchmark comprises a total of 3,800 manually curated and rigorously verified samples. As reflected in our experimental design, the dataset is strictly bifurcated into two evaluation tracks:
\begin{itemize}
    \item \textbf{Hierarchical Track (800 samples):} Dedicated to evaluating hierarchical topologies, exclusively utilizing nested JSON and XML formats to assess models' capabilities in managing recursive depth and long-range dependencies.
    \item \textbf{Tabular Track (3,000 samples):} Dedicated to evaluating grid-based topologies. This track covers CSV, HTML, Markdown, and LaTeX, alongside flattened JSON and XML lists, to systematically assess two-dimensional spatial alignment.
\end{itemize}

\subsection{System Prompt Configuration}
To minimize the variance introduced by prompt engineering and to rigorously evaluate the models' inherent zero-shot structured cognition, we employ a unified, deterministic system prompt across all models and formats. As implemented in our evaluation pipeline, the system prompt is strictly defined as follows:

\begin{quote}
\texttt{You are a data converter. Convert description into \{format\_type\}. Do not output any extra text or explanations. All the value should be stored in string type.}
\end{quote}

Here, \texttt{\{format\_type\}} is dynamically replaced by the target serialization format (e.g., \texttt{json}, \texttt{html}, \texttt{csv}) corresponding to the specific task. The explicit constraint ``\textit{All the value should be stored in string type}'' is enforced to prevent models from generating mixed data types, thereby isolating the evaluation to purely structural topology.

\subsection{User Input Generation Examples}
The input prompt (User Message) consists exclusively of the natural language description generated by the $\Phi$ function (introduced in Section \ref{sec:framework}). To avoid leaking serialization-specific syntax (such as braces \texttt{\{\}} or table pipes \texttt{|}), the descriptions are generated using algorithmic path-based and coordinate-based strategies. 

\subsubsection{\textbf{Example 1: Path-based Description (Hierarchical Track)}} 

For a hierarchical structure, the generator traverses the AST and produces strict path descriptions:
\begin{quote}
\texttt{Under the path root/users/id, there exists a node with value "101". Under the path root/users/name, there exists a node with value "Alice".}
\end{quote}

\subsubsection{\textbf{Example 2: Coordinate-based Description (Tabular Track)}} 

For a flat tabular structure, the generator reads the AST grid and produces spatial descriptions:
\begin{quote}
\texttt{The table has columns: "ID" and "Name". In row 1, the value of column "ID" is "101", and the value of column "Name" is "Alice".}
\end{quote}

By providing these pure semantic instructions, we guarantee that the Large Language Model must rely on its intrinsic understanding of data structures to reconstruct the target topology, effectively validating the rigor of our self-supervised evaluation loop.

% =========================================================
% Appendix B: 对应正文 4.1 节提到的完整实验数据 (后出现)
% =========================================================
\section{Comprehensive Results for ROUGE-1 and BLEU}
\label{app:experimental_results}

\begin{figure}[t]
    \centering
    \includegraphics[width=0.9\linewidth]{2_overall_performance_ROUGE-1.png}
    \caption{\textbf{Overall Performance Landscape and Metric Sensitivity.} The chart compares model performance across Hierarchical and Tabular tracks using CSA and NTED against the ROUGE-1 baseline.}
    \label{fig:a1_overall_performance_bar}

    \centering
    \includegraphics[width=0.9\linewidth]{2_overall_performance_BLEU.png}
    \caption{\textbf{Overall Performance Landscape and Metric Sensitivity.} The chart compares model performance across Hierarchical and Tabular tracks using CSA and NTED against the BLEU baseline.}
    \label{fig:a2_overall_performance_bar}
\end{figure}

Figure~\ref{fig:a1_overall_performance_bar}, Figure~\ref{fig:a2_overall_performance_bar}, and Table~\ref{tab:full_metrics} present the detailed performance landscape using traditional n-gram metrics (ROUGE-1, ROUGE-2, and BLEU) across both Hierarchical and Tabular tracks. These metrics serve as a comparative baseline to validate the sensitivity of our proposed framework.

\begin{table}[t]
    \centering
    \caption{\textbf{Full Performance Comparison on Traditional Metrics and Structure Scores.} We report ROUGE-1 (R-1), ROUGE-2 (R-2), BLEU, CSA, and NTED scores for both Hierarchical and Tabular tracks.}
    \label{tab:full_metrics}
    % 使用 resizebox 将宽度限制为 \linewidth，高度 ! 表示等比缩放
    \resizebox{\linewidth}{!}{
        \begin{tabular}{l|ccccc|ccccc}
            \toprule
            \multirow{2}{*}{\textbf{Model}} & \multicolumn{5}{c|}{\textbf{Hierarchical Track}} & \multicolumn{5}{c}{\textbf{Tabular Track}} \\
            \cmidrule(lr){2-6} \cmidrule(lr){7-11}
             & \textbf{R-1} & \textbf{R-2} & \textbf{BLEU} & \textbf{CSA} & \textbf{NTED} & \textbf{R-1} & \textbf{R-2} & \textbf{BLEU} & \textbf{CSA} & \textbf{NTED} \\
            \midrule
            DeepSeek-R1 & 0.8353 & 0.8146 & 0.3633 & 0.7386 & 0.7941 & 0.9698 & 0.9504 & 0.6249 & 0.9418 & 0.9746 \\
            DeepSeek-V3.1 & 0.9622 & 0.9259 & 0.5115 & 0.8351 & 0.9343 & 0.9624 & 0.9400 & 0.6400 & 0.9216 & 0.9759 \\
            DeepSeek-V3.2 & 0.9580 & 0.9167 & 0.5047 & 0.7999 & 0.9138 & 0.9623 & 0.9389 & 0.5852 & 0.8807 & 0.9741 \\
            GPT-OSS-120B & 0.9574 & 0.9127 & 0.5044 & 0.8491 & 0.9391 & 0.9707 & 0.9507 & 0.5617 & 0.9569 & 0.9813 \\
            Kimi-k2-thinking & 0.9477 & 0.9176 & 0.3535 & 0.8475 & 0.9194 & 0.9682 & 0.9521 & 0.6738 & 0.9574 & 0.9778 \\
            Llama-3.1-8B & 0.9369 & 0.8815 & 0.4655 & 0.6312 & 0.7962 & 0.9346 & 0.8845 & 0.5964 & 0.7543 & 0.9013 \\
            Llama-3.1-70B & 0.9708 & 0.9514 & 0.4564 & 0.8683 & 0.9351 & 0.9560 & 0.9220 & 0.6794 & 0.8820 & 0.9678 \\
            Minimax-M1-80k & 0.8697 & 0.8520 & 0.5076 & 0.7711 & 0.8194 & 0.9136 & 0.8979 & 0.6041 & 0.8593 & 0.9053 \\
            Ministral-8B & 0.9441 & 0.9000 & 0.4867 & 0.6950 & 0.7867 & 0.9550 & 0.9282 & 0.6912 & 0.8812 & 0.9411 \\
            Qwen2.5-72B & 0.9697 & 0.9367 & 0.5129 & 0.8447 & 0.9303 & 0.9711 & 0.9510 & 0.6193 & 0.8893 & 0.9638 \\
            Qwen3-0.6B & 0.5060 & 0.4417 & 0.2491 & 0.1136 & 0.2106 & 0.4923 & 0.4401 & 0.1524 & 0.1726 & 0.4063 \\
            Qwen3-32B & 0.9685 & 0.9351 & 0.5156 & 0.8125 & 0.9157 & 0.9654 & 0.9412 & 0.6105 & 0.9655 & 0.9804 \\
            Qwen3-235B & 0.9231 & 0.8909 & 0.4686 & 0.7829 & 0.8708 & 0.9619 & 0.9433 & 0.5749 & 0.9338 & 0.9593 \\
            \bottomrule
        \end{tabular}
    } % 结束 resizebox
\end{table}

A critical insight from these results is the phenomenon of \textbf{discriminative saturation} observed in traditional metrics. As detailed in Section \ref{subsec:performance_overview}, ROUGE and BLEU scores tend to cluster within a narrow range with low statistical variance ($\sigma^2_{BLEU} \approx 0.0056$), often failing to penalize structurally invalid outputs that retain high surface-level lexical overlapping. In contrast, CSA and NTED metrics exhibit significantly higher variance ($\sigma^2_{CSA} \approx 0.0601$), effectively amplifying the performance gaps between models. This heightened sensitivity confirms that CSA and NTED possess superior \textbf{discriminative power}, enabling a more rigorous diagnosis of a model's ability to maintain topological integrity versus merely mimicking text patterns.

\section{Fine-Grained Performance Heatmaps}
\label{app:heatmap_analysis}

\begin{figure}[h]
    \centering
    \includegraphics[width=1.0\textwidth]{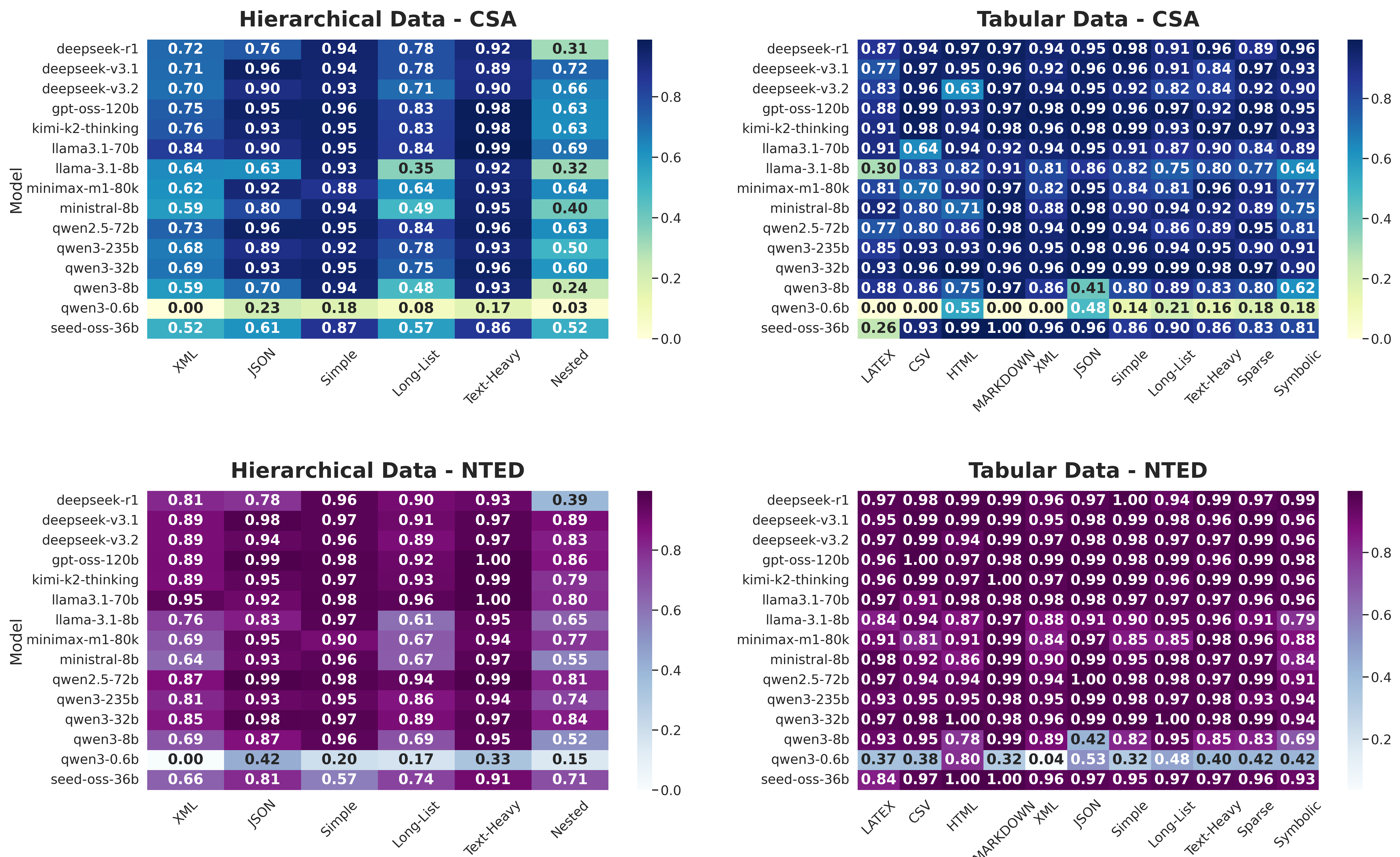}
    \caption{\textbf{Comprehensive Performance Heatmap.} The top row displays CSA, reflecting the fidelity of information recovery. The bottom row displays NTED, measuring structural topological integrity. Darker blue (CSA) and purple (NTED) indicate superior performance, while lighter shades reveal performance bottlenecks.}
    \label{fig:comprehensive_heatmap}
\end{figure}

We present a comprehensive heatmap analysis in Figure \ref{fig:comprehensive_heatmap} to provide a granular examination of model capabilities across heterogeneous topologies. This visualization decomposes the performance of all 15 evaluated models according to specific data formats and complexity characteristics.

\subsection{Hierarchical Track Analysis}
The analysis of Hierarchical Data reveals significant variances in topological adaptability. A primary observation is the consistent performance bottleneck presented by nested tasks, as discussed in Section \ref{subsec:complexity_robustness}. This is particularly evident in the reasoning-enhanced DeepSeek-R1, which exhibits a sharp performance divergence; while it achieves a high CSA of 0.94 on Simple tasks, its performance declines precipitously to 0.31 on Nested tasks. This visual evidence supports the hypothesis that current Chain-of-Thought mechanisms may prioritize symbolic manipulation at the expense of recursive memory stack maintenance. Furthermore, a comparison between XML and JSON columns indicates a widespread inductive bias towards JSON. For instance, Minimax-M1-80k achieves a CSA of 0.92 on JSON but drops significantly to 0.62 on XML. This disparity likely reflects a code-native bias resulting from the dominance of code data in pre-training corpora. Additionally, the smallest model, Qwen3-0.6B, exhibits near-zero scores across most hierarchical tasks, illustrating the phenomenon of hollow structural generation where the model fails to construct even basic valid topologies.

\subsection{Tabular Track Analysis}
In the Tabular Track, the verbosity of syntax exerts a notable influence on performance. While most models demonstrate robust capabilities on Markdown and CSV formats, the HTML format triggers distinct stratification. Despite being a large-scale model, DeepSeek-V3.2 experiences a significant performance drop on HTML (CSA: 0.63) compared to its performance on Markdown (CSA: 0.97). In contrast, other large-scale models such as Qwen3-235B (CSA: 0.93) and Llama-3.1-70B (CSA: 0.94) maintain high fidelity, suggesting that specific architectural optimization or training data distribution, rather than parameter scale alone, determines the ability to suppress noise introduced by redundant HTML tags. Moreover, the Qwen3-32B model demonstrates remarkable cost-efficiency and stability. It maintains high scores across diverse formats (e.g., CSA $>$ 0.90 for both JSON and Markdown), rivalling the performance of massive models like DeepSeek-V3.1 and Qwen3-235B.

\subsection{Correlation Between Content and Structure}
A comparative analysis of the top row (CSA) and bottom row (NTED) reveals a strong positive correlation between content fidelity and structural integrity. However, it is observed that NTED scores generally surpass CSA scores in weaker models. This discrepancy indicates a learning trajectory where models acquire the syntactic skeleton of structured data before mastering the precise semantic content, thereby validating the necessity of our two-stage metric design.

\section{Case Study}
This section presents representative failure cases to elucidate a recurring phenomenon observed throughout our experiments: high surface-form similarity can systematically mask fatal structural violations. We analyze three archetypal failure modes across flat and hierarchical generation settings, demonstrating how conventional n-gram metrics remain insensitive to these errors, while CSA and NTED respond appropriately.

\paragraph{Case 1: Column Cardinality Violation (Flat Structures).}
Table~\ref{fig:csv_failure_case} illustrates a common failure in CSV generation, where an escaped numeric value (\texttt{``20,590.90''}) is incorrectly interpreted as a delimiter. This error silently expands the column cardinality, shifting all subsequent fields. Despite this structural corruption, surface-based similarity remains high (BLEU = 0.9571), as most tokens are preserved. In contrast, CSA sharply degrades (0.5660), correctly reflecting the violation of the table schema. This case highlights that token overlap alone is insufficient to capture structural alignment in flat data representations.

\begin{table}[htbp] % 1. 将 table* 改为 table
    \centering
    \small
    \setlength{\tabcolsep}{4pt}
    \begin{tabular}{lcccccc}
        \toprule
        \textbf{Source} & \textbf{Month} & \textbf{Price} & \textbf{Balance} & \textbf{Desc.} & \textbf{RawLog} & \textbf{\textcolor{red}{Extra Col.}} \\
        \midrule
        \textbf{Ground Truth} & 05 & 262.75 & \textbf{20,590.90} & \textit{<Desc>} & \textit{<Raw>} & \textit{--} \\
        \cmidrule(lr){1-7}
        \textbf{Model Output} & 05 & 262.75 & \textbf{20} & \textbf{590.90} & \textit{<Desc>} & \textit{\textcolor{red}{<Raw>}} \\
        \bottomrule
        \addlinespace
        % 2. 核心修复：将 c 改为 p{...} 格式，允许自动换行，并在内部使用 \centering 保持居中
        \multicolumn{7}{p{0.95\linewidth}}{
            \centering
            \textbf{Metric Discrepancy:} 
            BLEU: \textbf{0.9571} (High Similarity) vs. 
            CSA: \textbf{0.5660} (Structural Failure)
        } \\
    \end{tabular}
    \caption{\textbf{Case 1: Column Cardinality Violation.} The model incorrectly splits the comma-separated numeric value `20,590.90` into two separate fields (`20` and `590.90`). This introduces an erroneous 6th column, causing a severe structural misalignment (high NTED/low CSA) despite high surface-level textual overlap (high BLEU).}
    \label{fig:csv_failure_case}
\end{table}

\paragraph{Case 2: State Confusion in Hierarchical Generation.}
As shown in Table~\ref{tab:json_compound_failure}, the model exhibits a compound failure in JSON generation by losing track of the active schema state. Specifically, it (i) omits a required comma between list elements, (ii) injects a key--value pair into an array context, and (iii) mismatches closing delimiters by terminating an array scope with an object brace. While the output remains locally plausible at the token level, the global structure becomes invalid. This failure exemplifies state confusion, where the model fails to maintain hierarchical context across generation steps.

\begin{table}[htbp] % 单栏环境改用 table 即可
    \centering
    \small
    % 将 0.48 改为 0.45，留出足够的 \tabcolsep 空间
    \begin{tabular}{p{0.45\linewidth} p{0.45\linewidth}}
        \toprule
        \textbf{Ground Truth (Valid)} & \textbf{Model Output (Compound Failure)} \\
        \midrule
        \begin{minipage}[t]{\linewidth}
        \ttfamily\small
        \dots \\
        "subtasks": \textbf{\textcolor{blue}{[}} \\
        \hspace*{4mm}\{ \\
        \hspace*{8mm}"id": "101", \\
        \hspace*{8mm}"hours": "5" \\
        \hspace*{4mm}\}\textbf{\textcolor{blue}{,}} \textit{// Comma} \\
        \hspace*{4mm}\{ \\
        \hspace*{8mm}"id": "102", \\
        \dots \\
        \hspace*{4mm}\} \\
        \textbf{\textcolor{blue}{]}}, \textit{// Array match} \\
        \textbf{"taskId": "999"} \textit{// Sibling} \\
        \dots
        \end{minipage}
        & 
        \begin{minipage}[t]{\linewidth}
        \ttfamily\small
        \dots \\
        "subtasks": \textbf{\textcolor{blue}{[}} \textit{// Open Array} \\
        \hspace*{4mm}\{ \\
        \hspace*{8mm}"id": "101", \\
        \hspace*{8mm}"hours": "5" \\
        \hspace*{4mm}\} \textbf{\textcolor{red}{\fbox{?}}} \textit{\scriptsize No Comma} \\
        \vspace{1mm}
        % Error: Key-Value in List
        \hspace*{4mm}\textbf{\textcolor{red}{"taskId": "999"}} \textit{// Invalid Item} \\
        \vspace{1mm}
        % Error: Bracket Mismatch
        \textbf{\textcolor{red}{\}}} \textbf{\textcolor{red}{\fbox{?}}} \textit{\scriptsize Expected `]`} \\
        \dots \\
        \end{minipage} \\
        \bottomrule
        \addlinespace
        % 关键修复：将 c 改为 p{...}，使其能够自动折行
        \multicolumn{2}{p{0.95\linewidth}}{
            \textbf{Parser Exception:} 
            \texttt{Expecting ',' delimiter} \textit{AND} \texttt{Mismatched brackets: '[' closed by '\char`\}'}
        } \\
    \end{tabular}
    \caption{\textbf{Case 2: Compound State Confusion.} The model loses track of the schema context entirely.}
    \label{tab:json_compound_failure}
\end{table}

\paragraph{Case 3: Syntax Fracture during Deep-Nesting Transitions.}
Table~\ref{tab:syntax_fix} presents a minimal yet illustrative example of a transition failure in deeply nested JSON. After correctly completing the \texttt{cloudInfrastructure} object, the model fails to emit the required delimiter before continuing with a sibling field. Such errors are catastrophic for downstream parsing, yet remain weakly penalized by surface similarity metrics.

Across all cases, the generated outputs preserve substantial lexical overlap with the ground truth, leading to deceptively high BLEU scores. Nevertheless, each example contains a structural defect that renders the output semantically or syntactically invalid. These findings reinforce the necessity of topology-aware evaluation: CSA and NTED expose failure modes that remain invisible to surface-form metrics, providing a more faithful assessment of structured generation fidelity.

\begin{table}[htbp]
    \centering
    \small
    \begin{tabular}{|p{0.45\linewidth}|p{0.45\linewidth}|}
        \hline
        \textbf{Ground Truth (Valid)} & \textbf{Model Output (Syntax Error)} \\
        \hline
        \begin{minipage}[t]{\linewidth}
            \vspace{1mm} 
            \ttfamily\scriptsize 
            \dots \\
            "cloudInfrastructure": \{ \\
            \hspace*{4mm}"aws": "12345", \\
            \hspace*{4mm}"gcp": "12345" \\
            \}\textbf{\textcolor{blue}{,}} \textit{// Comma Required} \\
            \vspace{1mm}
            \textbf{"employeeBenefits": \{} \\
            \hspace*{4mm}"health": "12345" \\
            \dots
            \vspace{1mm} 
        \end{minipage}
        & 
        \begin{minipage}[t]{\linewidth}
            \vspace{1mm}
            \ttfamily\scriptsize
            \dots \\
            "cloudInfrastructure": \{ \\
            \hspace*{4mm}"aws": "12345", \\
            \hspace*{4mm}"gcp": "12345" \\
            \} \textbf{\textcolor{red}{\fbox{?}}} \textit{Missing comma} \\
            \vspace{1mm}
            \textbf{"employeeBenefits": \{} \\
            \hspace*{4mm}"health": "12345" \\
            \dots
            \vspace{1mm}
        \end{minipage} \\
        \hline
        % 关键修复：直接使用 p 列，无需再嵌套一层 minipage
        \multicolumn{2}{|p{\dimexpr 0.9\linewidth + 2\tabcolsep \relax}|}{
            \vspace{1mm}
            \centering
            \textbf{Failure Diagnosis:} \\
            % 如果错误日志极长，可以在标点符号后人为加一个空格允许换行，或使用 \url{} 的方式
            \textit{Error Log:} \texttt{Expecting ',' delimiter: line 117...} \\
            The model closed the object scope but failed to predict the transition token (comma).
            \vspace{1mm}
        } \\
        \hline
    \end{tabular}
    \caption{\textbf{Case 3: Syntax Fracture Case.} The model omits the trailing comma after closing the \texttt{cloudInfrastructure} object, causing a fatal parsing error.}
    \label{tab:syntax_fix}
\end{table}

\end{document}